\documentclass[10pt,twocolumn,letterpaper]{article}

\usepackage[pagenumbers]{cvpr} 

\usepackage{graphicx}
\usepackage{amsmath}
\usepackage{amssymb}
\usepackage{booktabs}

\usepackage[accsupp]{axessibility}

\usepackage[pagebackref,breaklinks,colorlinks]{hyperref}

\usepackage[capitalize]{cleveref}
\crefname{section}{Sec.}{Secs.}
\Crefname{section}{Section}{Sections}
\Crefname{table}{Table}{Tables}
\crefname{table}{Tab.}{Tabs.}

\def\cvprPaperID{19} 
\def\confName{CVPRW}
\def\confYear{2023}

\usepackage[dvipsnames]{xcolor}

\begin{document}

\title{What Affects Learned Equivariance in Deep Image Recognition Models?

\vspace{3mm}}

\author{
Robert-Jan Bruintjes$^{*,1}$, Tomasz Motyka\thanks{Equal contribution.}$\hspace{1.5mm}$$^{,2}$, Jan van Gemert$^{1}$\\
$^{1}$Computer Vision Lab, Delft University of Technology, $^{2}$Synerise\\
{\tt\small \{r.bruintjes,j.c.vangemert\}@tudelft.nl, tomasz.motyka@synerise.com}
}
\maketitle

\begin{abstract}

Equivariance w.r.t. geometric transformations in neural networks improves data efficiency, parameter efficiency and robustness to out-of-domain perspective shifts. When equivariance is not designed into a neural network, the network can still learn equivariant functions from the data. We quantify this learned equivariance, by proposing an improved measure for equivariance. We find evidence for a correlation between learned translation equivariance and validation accuracy on ImageNet. We therefore investigate what can increase the learned equivariance in neural networks, and find that data augmentation, reduced model capacity and inductive bias in the form of convolutions induce higher learned equivariance in neural networks.

\end{abstract}

\section{Introduction}

\begin{figure}[t]
    \centering
    \includegraphics[width=1.0\linewidth]{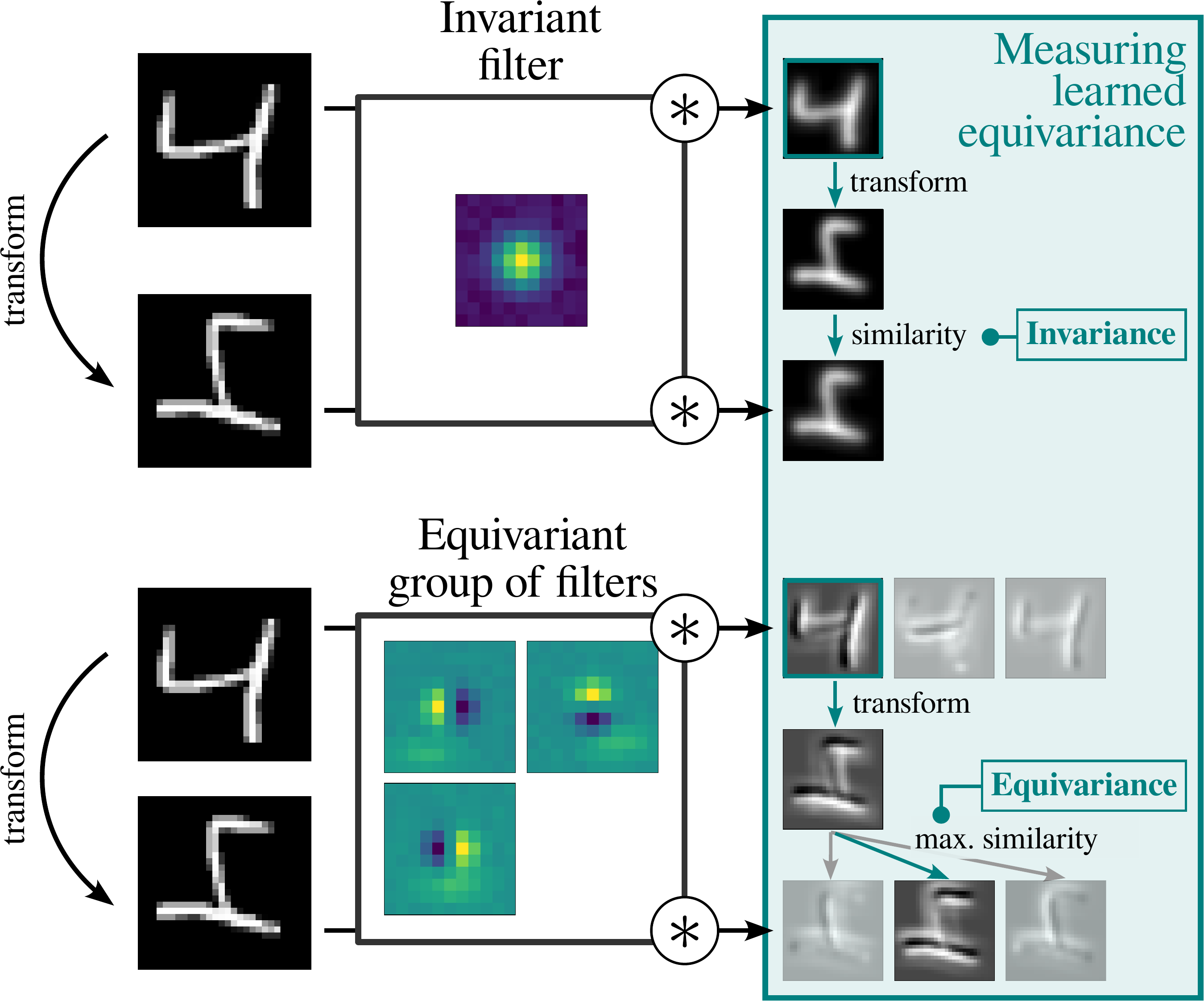}
    \caption{Neural networks can learn features that are invariant or equivariant w.r.t. a geometric transformation of the data, such as rotation. We measure learned equivariance w.r.t. translation and rotation in neural networks.}
    \label{fig:1}
\end{figure}

Equivariance in neural network features allows invariance to geometric transformations \cite{cnn, gconv}, making such networks more data efficient \cite{steerable_equivariance, winkels_3d_2018, lengyel_exploiting_2021}, parameter efficient \cite{gconv} and robust to out-of-distribution transformations~\cite{altstidl2022just, edinxhoventom2023, sangalli2022scale}. 

Equivariance with respect to specific geometric transformations can be designed into the neural network architecture~\cite{gconv, e-2_steerable, cohen2019gauge}. However, even with careful design, it may happen that the resulting architecture is not as equivariant as intended \cite{edinxhoventom2023, blurpool, kayhan_translation_2020, pos_in_channel}.
An example is the convolution operator in Convolutional Neural Networks (CNNs) for translation equivariance, which can be broken by border effects \cite{kayhan_translation_2020, pos_in_channel} or pooling~\cite{blurpool}. 
On the other hand, even if  neural networks are not designed to be equivariant, they can still \textit{learn} equivariance naturally. Existing works demonstrate qualitative examples of learned equivariant features \cite{bergstra_statistical_2011, dieleman_exploiting_2016, olah_naturally_2020}. However, how much equivariance is learned, and which factors affect equivariance, are open questions. 

In this work, we quantify learned equivariance in image recognition neural networks that have and have not been explicitly designed for equivariance. Where existing works~\cite{inv_belief, inv_transfer, quantifying_invariance, biscione2020learning, tracking_invariance} typically only measure equivariance at the output of the network, we measure equivariance for all intermediate layers. To do so, we deviate from existing measures of learned equivariance which are inconsistent across network depths, and we design a consistent measure.

Using our measure for learned equivariance, we find evidence that learned translation equivariance in intermediate features of neural networks correlates with increased validation accuracy on ImageNet.
We therefore investigate how we can increase learned equivariance by changing how we train neural networks. In particular, we find that 1) making the task equivariant does not increased learned equivariance; 2) data augmentations designed for invariance indeed increase learned equivariance, even in early and middle layers; 3) reducing model capacity increases learned equivariance, suggesting that equivariant features arise from a need to compress representations; 4) CNNs learn more translation and rotation equivariance in intermediate features than the Vision Transformers (ViTs).

We make the following contributions:

\begin{itemize}
    \item We propose a new measure for learned equivariance that is allows comparing learned equivariance of features at different depths of the network.
    \item We show evidence for a positive correlation between learned translation equivariance in intermediate features and validation accuracy on ImageNet.
    \item We test how several aspects of neural network training affect learned equivariance. In summary, we find that data augmentation, reduced model capacity and the inductive biases of CNNs positively affect learned equivariance.
\end{itemize}

\section{Related Works}

Neural networks can learn equivariant features from data \cite{understanding_represenations,early_vision,olah_naturally_2020}. Particularly inspiring is the work by Olah \etal~\cite{olah_naturally_2020}, that demonstrates by precise and meticulous manual investigations that learned equivariant features exist in networks that were not designed to be equivariant. Inspired by this work, we here investigate how to move beyond laborous manual qualitative investigations, and instead offer a quantitative approach, by giving an automatic measure for learned equivariance.  

A number of existing works measure equivariance in neural networks. \cite{inv_belief} study models from the pre-Deep Learning era which have since been superseded by the models we study. More recent works measure equivariance in Convolutional Neural Networks, with KL divergence on class probabilities~\cite{inv_transfer}, with Euclidean distance~\cite{quantifying_invariance}, or cosine similarity~\cite{biscione2020learning, tracking_invariance} on feature maps. 
In our work, we show that the cosine similarity is not appropriate for measuring equivariance in intermediate feature maps, and offer a correlation-based measure.

Several works study how neural network hyperparameters and datasets affect learned translation equivariance in the final output of the network. The kernel and padding sizes of the architecture affect translation invariance~\cite{tracking_invariance}, although data augmentation might have a bigger effect on translation invariance than the network architecture~\cite{quantifying_invariance}. Similar conclusions about the importance of the data were found by others~\cite{biscione2020learning, grounding_invariances}.
Here, we follow these investigations, and extend them by analyzing the impact on the intermediate layers.

There are some works that study equivariant properties of intermediate features. Recently, \cite{moskalev2022liegg} proposed a method to detect invariance to any learned Lie group for intermediate features. However, they do not study equivariance, like we do. Other works 
study only the transformation group of translations ($\mathbb{Z}^2$). \cite{blurpool} measures the translation equivariance by computing cosine similarity between feature maps to show how max pooling violates the translation equivariance property. \cite{kayhan_translation_2020,pos_in_channel} show that some padding methods disrupt the translation equivariance property in CNNs. \cite{pooling_stability} measure the invariance of intermediate representations using normalized cosine similarity to study the effect of pooling on deformation stability. Where these works diagnose issues with designed equivariance and test for their effects, we consider learned equivariance in a more general sense, including transformation groups not designed into the network, such as rotations.


\section{Method}

Neural networks can learn to be equivariant in two ways: either by learning invariant features or by learning equivariant groups of features, as shown in Fig~\ref{fig:1}. In this section we detail how we can measure the quantity of invariant features and equivariant groups of features. We discuss which similarity measure is appropriate for measurements of learned equivariance in features at different depths of a neural network. Finally, we verify our measures using artificially engineered equivariant CNNs.

In the following we will refer to invariant features and equivariant groups of features under the single predicate "learned equivariance", as invariance is a special case of equivariance.

\subsection{Invariant features}

We derive a measure of learned equivariance from inspecting the definition of equivariance \cite{gconv} applied to a single neural network layer:

\begin{equation}
    \label{eq:equivariance}
    f(T_g(X)) = T_g'(f(X)),
\end{equation}

where $X \in \mathbb{R}^{C_{\text{in}} \times H \times W}$ is an image or a feature map, $f(X) \in \mathbb{R}^{C \times H \times W}$ is the output of a neural network operation with $C$ output features and $T_g$ is the application of a transformation $g$ from a transformation group $G$. For example, if $G = \mathbb{Z}^2$, then $g$ is a translation with a particular integer-valued $(x,y)$ offset. If $T_g'$ is the identity function for all $g$, the layer $f$ is \textit{invariant} w.r.t. transformation $T$:

\begin{equation}
    f(T(X)) = f(X).
\end{equation}

Without designing invariance to $T$ into neural network layer $f$, each individual feature in $f(X)$ can learn to behave invariant or not invariant with respect to $T$. We therefore define invariance for each feature $c \in C$ independently:

\begin{equation}
    f(T(X))_{c} = f(X)_{c}.
\end{equation}

In Fig.~\ref{fig:1} we show an example where $T$ is a $90^{\circ}$ rotation. 

To measure a feature's invariance w.r.t. $g$, we compute the similarity between $f(T(X))_{c}$ and $f(X)_{c}$:

\begin{equation}
    \text{Invariance}(f_c,g) = S(f(T(X))_{c}, f(X)_{c}),
\end{equation}

given a similarity function $S : \mathbb{R^{H \times W}} \times \mathbb{R^{H \times W}} \to [0, 1]$. Given invariance measures for each feature, we can average these measures for all features in a layer to compute a layer's invariance.


\subsection{Equivariant features}

\begin{figure*}[t]
     \centering
     \begin{subfigure}{0.48\textwidth}
         \centering
         \includegraphics[width=0.8\textwidth]{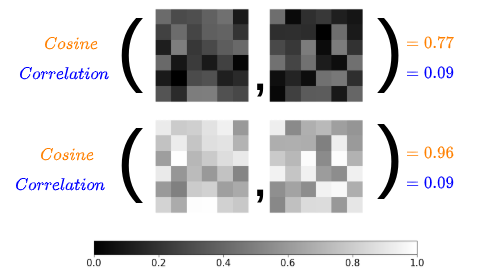}
         \caption{Comparison of different similarity measures on random feature maps. Feature maps on the bottom are shifted by 0.5 with respect to the top ones. Cosine similarity is sensitive to such shifts while correlation is invariant.}
         \label{fig:feature_map_cosine}
     \end{subfigure}
     \hfill
     \begin{subfigure}{0.48\textwidth}
         \centering
         \includegraphics[width=\textwidth]{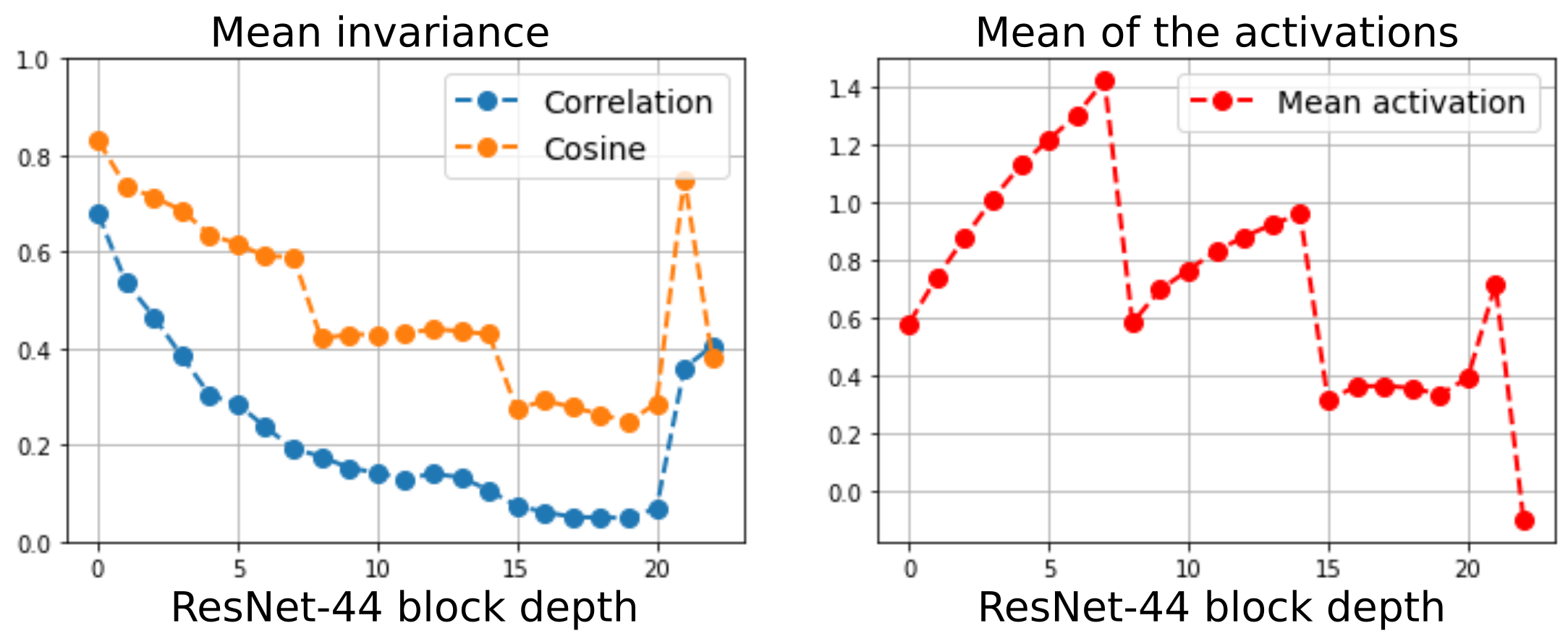}
         \caption{Comparison between magnitude of the activations (red) and equivariance computed using correlation (blue) and cosine similarity (orange). We compute correlation between the magnitude of the activations and equivariance scores computed with correlation (\textbf{0.11}) and cosine similarity (\textbf{0.63}).}
         \label{fig:mean_bias_depth}
     \end{subfigure}
 \caption{Analysis of the influence of magnitude of weights on different similarity measures. }
 \label{fig:mean_bias}
\end{figure*}

A group of features $C_G \subseteq C$ in a neural network layer is equivariant with respect to a transformation group $G$ if each feature $c' \in C_G$ activates for a different transformation $g$ from the group. In other words, for a sample $T_g(X)$ transformed with any transformation from the group, the group of feature maps $f(X)_{c' \in C_G}$ will have one feature $c'$ whose transformed feature map $T_g(f(X))_{c'}$ matches $f(T_g(X))_{c'}$:



\begin{equation}
    f(T_g(X))_{c'} = T_g(f(X))_{c'}, \quad \exists c' \in C_T.
\end{equation}

In Fig. \ref{fig:1} we show an example where $g$ is a $90^{\circ}$ rotation.

To fit this with the definition of equivariance (Eq.~\ref{eq:equivariance}) we define $T'_g$:

\begin{equation}
    f(T_g(X))_{c' \in C_G} = T'_g(f(X))_{c' \in C_G}
\end{equation}

where $T'_g$ transforms with $T_g$ and selects feature $c'$ that matches the transformation $g$. When this equation holds, the feature group $C_G$ is equivariant w.r.t. $T$.

To measure the equivariance of a feature $c$ we find the maximum similarity between $f(T_g(x))_{c}$ and $T_g(f(x))_{c'}$ over all features $c' \in C$, for a given transformation $g$:

\begin{align}
    \text{Equivariance}(f_c,g) &= \max_{c' \in C} S(f(T_g(X))_{c}, T_g(f(X))_{c'})
\end{align}

given a similarity function $S : \mathbb{R^{H \times W}} \times \mathbb{R^{H \times W}} \to [0, 1]$. Given equivariance measures for each feature, we can average these measures for all features in a layer to compute a layer's equivariance. Note that if a feature is invariant w.r.t. $T_g$, we will measure an equivariance score that is at least as high as the invariance score. As invariance is a special case of equivariance, this behavior of our measure is intended.

\subsection{Measuring similarity}
\label{sec:similarity}

\begin{figure*}[t]
     \centering
     \begin{subfigure}{0.48\textwidth}
         \centering
         \includegraphics[width=\textwidth]{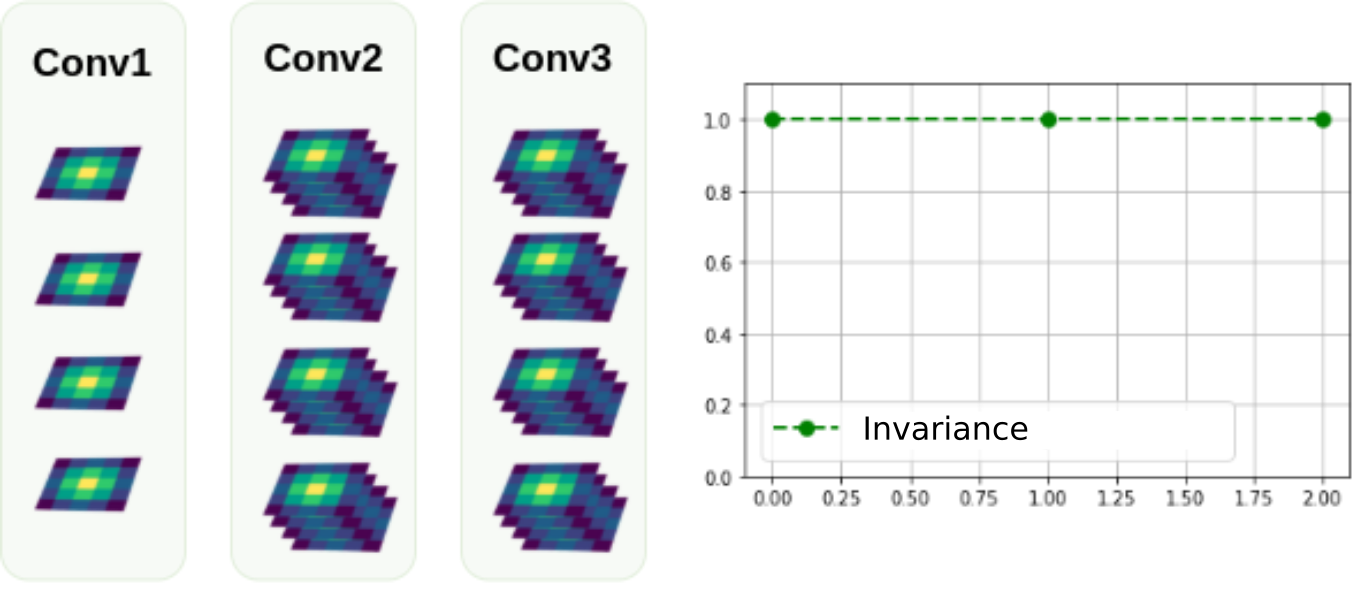}
         \caption{Rotation invariance}
         \label{fig:perfect_channel}
     \end{subfigure}
     \hfill
     \begin{subfigure}{0.48\textwidth}
         \centering
         \includegraphics[width=\textwidth]{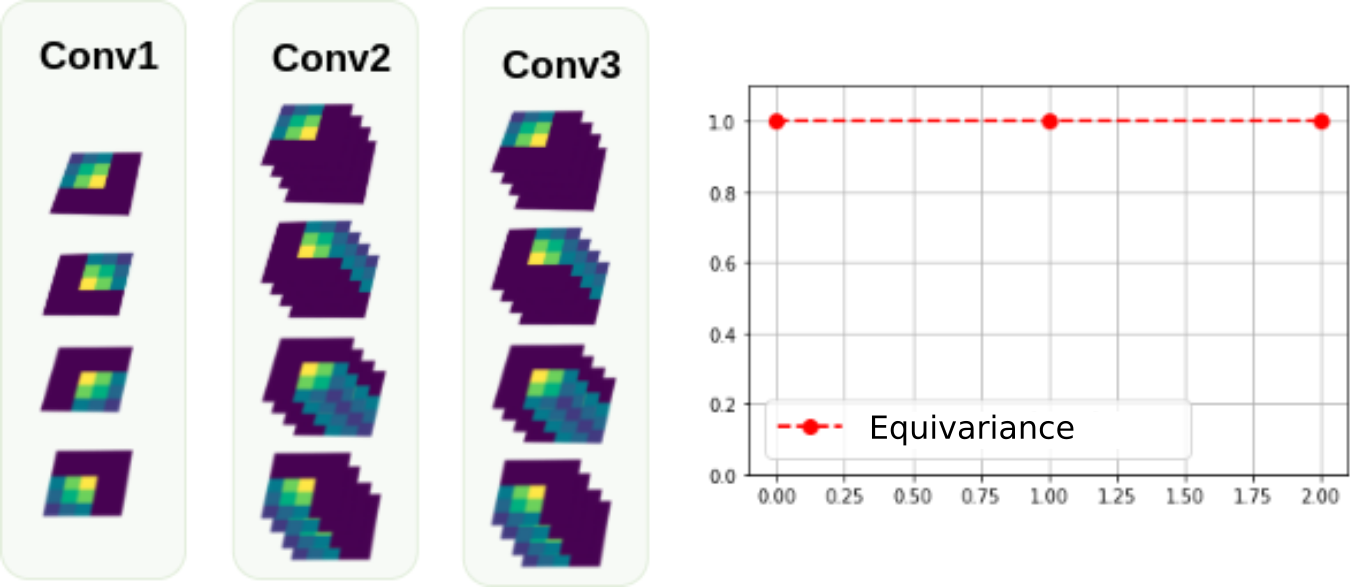}
         \caption{Rotation equivariance}
         \label{fig:perfect_layer}
     \end{subfigure}
     \caption{We create two controlled toy CNNs, each designed to be perfectly invariance and equivariant respectively, to test if our method measures equivariance correctly, which it does.}
     \label{fig:type_1}
\end{figure*}

We need to choose a similarity measure $S : \mathbb{R^{H \times W}} \times \mathbb{R^{H \times W}} \to [0, 1]$ with which to compare feature maps to measure invariance and equivariance. In existing works, cosine similarity is commonly used as a similarity measure used to compute invariance or equivariance of the networks representations \cite{grounding_invariances,tracking_invariance,blurpool}. However, cosine similarity is sensitive to the mean values of its input vectors. This behaviour is depicted in Figure \ref{fig:feature_map_cosine}. As different layers in a neural network have different mean activation values (see Fig.~\ref{fig:mean_bias_depth}), this biases the similarity measure.

We propose to use Pearson correlation \cite{correlation} instead. Pearson correlation, also called centered cosine similarity, is a similarity measure that does not suffer from sensitivity to the mean of the inputs, as it computes the covariance of the inputs normalized by their standard deviations. It is the basis of many methods for comparing network representations \cite{cka,cca,svcca}.


To motivate our choice, we visualize the difference between using cosine similarity and correlation for measuring equivariance in the following example. We train ResNet-44 \cite{resnet} model on CIFAR-10 \cite{cifar10} dataset and compute invariance w.r.t. to $90^{\circ}$ rotation after each residual block. In Figure \ref{fig:mean_bias} we show the qualitative comparison between the scores, computed using cosine similarity and correlation, and the mean of the activations. Additionally, we compute a correlation between the magnitude of the activations and equivariance scores computed with 
cosine similarity (\textbf{0.63}) and correlation (\textbf{0.11}). Scores computed using cosine similarity correlate visibly with the mean of the activation while, for the scores computed using correlation, this effect is less prevalent. In our experiments, we therefore use correlation as a measure to quantify equivariance.

\section{Experiments}

\subsection{Controlled experiments}

To verify that our method captures equivariance, we apply it to two controlled toy settings. We create two 3-layer CNNs with hand-crafted filters such that we expect to measure perfect learned rotation invariance and equivariance respectively.
For the invariant model, we set all the filters to be rotationally symmetric, using a 2D isotropic Gaussian function, and measure the invariance after each layer (Fig. \ref{fig:perfect_channel}). For the equivariant model, we cut out corners of the filters from the invariant model such that all the filters are rotations of one another (Fig. \ref{fig:perfect_layer}). Our measure finds both models capture exactly the intended learned equivariances, demonstrating the validity of our measure.

\subsection{Does learned equivariance improve accuracy?}
\label{sec:correlation_acc}
We study the relationship between the validation accuracy and the amount of learned equivariance in large-scale seminal models. For each part of each trained model we compute Spearman's rank correlation between the amount of invariance or equivariance and the ImageNet validation accuracy of the model.

We test four CNNs (EfficientNet-B6 \& EfficientNet-B7 \cite{efficient_net}, ResNeXT-101 \cite{resnext} and Inception-V3 \cite{inception_v3}) and two Vision Transformer variants (Vision Transformer \cite{vit} and MLP-Mixer \cite{mlpmixer}). We measure invariance and equivariance for both translation and rotation for 2000 images from the ImageNet validation set. We do not train the models ourselves but instead use available checkpoints from \texttt{torchvison} \cite{pytorch} or \texttt{timm} \cite{timm}. Since the studied model families do not have the same exact number of layers, we divide each model into depth-wise parts and report the average equivariance measures over all layers in each part. Feature maps from the beginning of the network until the global average pooling (GAP) layer are uniformly partitioned into \textit{Early}, \textit{Middle} and \textit{Late} parts. \textit{Pool} captures the feature maps directly after the GAP layer and \textit{Final} is the feature map directly before the softmax layer. We discriminate between the \textit{Pool} part and the \textit{Final} part to identify what role in achieving equivariance the global pooling and final classifier have.

\begin{figure}[t]
    \centering
    \includegraphics[width=1.0\linewidth]{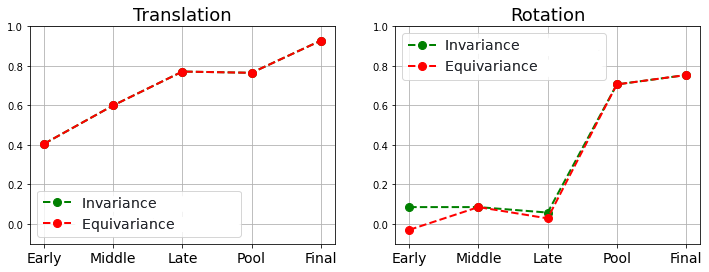}
    \caption{Spearman's rank correlation between learned equivariance and ImageNet validation accuracy. Translation equivariance in intermediate features correlates with increased accuracy on ImageNet, while rotation equivariance does not.}
    \label{fig:eq_acc}
\end{figure}

Figure \ref{fig:eq_acc} shows there is some correlation between translation equivariance in \textit{Early} and \textit{Middle} layers and accuracy on ImageNet, while attaining almost perfect correlation in the \textit{Final} part. In contrast, for rotations there is little correlation between the equivariance in the representation before global pooling and the validation accuracy.

Even though the sample size (six models) for this correlation test is small, we conclude that there is some evidence for the benefit of learning translation equivariance in intermediate features of neural networks trained on ImageNet. In the following we therefore study what can increase the learned equivariance in such networks.

\subsection{Equivariance in the data}
\label{sec:symmetries}

On tasks where invariant responses are beneficial to solve the task, e.g. translation invariance in image recognition, one may wonder how this invariance is achieved. We study how learned equivariance in intermediate features is affected by adding transformations to the data and therefore into the task. We choose to study rotation transformations on CNNs, as rotation equivariance is not designed into CNNs. We study whether there is a difference if the task is invariant or equivariant with respect to introduced transformations.

We train a 7-layer CNN taken from \cite{gconv}, consisting of 7 layers of $3 \times 3$ convolutions, 20 channels in each layer, ReLU activation functions, batch normalization, and max-pooling after layer 2, on 3 different datasets. We test on three different datasets. The first dataset is \textit{MNIST6}, which is the regular MNIST\cite{mnist} without $\{0, 1, 6, 8\}$ classes, to get rid of rotational transformations that these classes have. For example, digit 8 is very similar to its $180^{\circ}$ rotation, so, by default, this class would introduce some rotation invariance, which is undesirable as we want to control for rotation invariance in this setting. Second is the \textit{MNIST6-Rot-Inv} where every digit in \textit{MNIST6} is randomly rotated by $r \in \{0^{\circ}, 90^{\circ}, 180^{\circ}, 270^{\circ}\}$ upfront. This dataset imposes invariance into the task as, for every transformation, the predicted class should be the same. The last dataset, \textit{MNIST6-Rot-Eq}, is created in the same way as \textit{MNIST6-Rot-Inv}, but now the classes are made up of all combinations of digit number and rotation (e.g. class 2 is $(\text{digit} 0^{\circ}, 180^{\circ})$). This dataset imposes equivariance into the task. We compute in- and equivariance of the trained model for 2000 images from the validation set. We average the score over ${90^{\circ}, 180^{\circ}, 270^{\circ}}$ rotations. Each experiment is repeated three times using different random seed. We train for 100 epochs using Adam with a batch size of 128 and a learning rate of 0.01, L2 regularization at 0.0005 and weight decay at epochs 25 and 50 with a factor of 10.0.

\begin{figure}[t]
    \centering
    \includegraphics[width=1.0\linewidth]{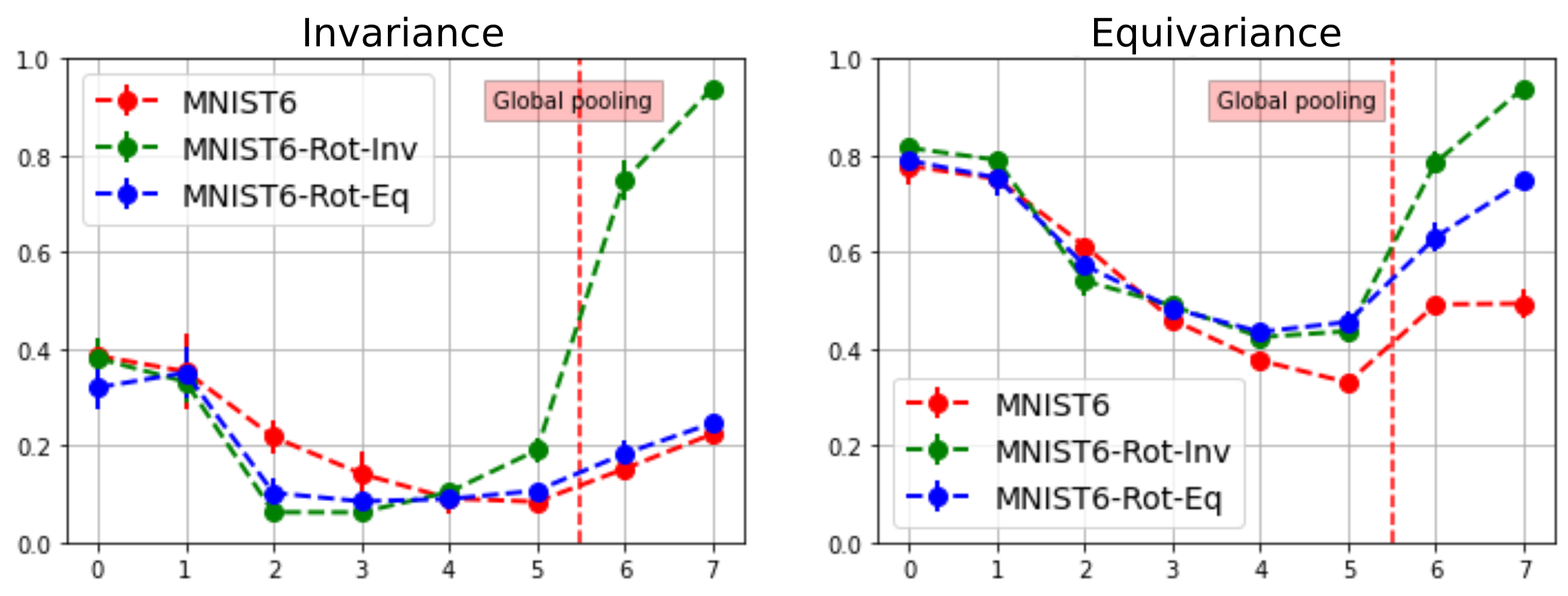}
    \caption{Learned rotation equivariance for rotation invariance/equivariance in the data. Invariance or equivariance in the task does not induce learning more equivariant features up until the late part of the network. Also there is no visible difference, up until the late part of the network, in the learned equivariance between the invariant and equivariant tasks.}
    \label{fig:objective}
\end{figure}

In Figure \ref{fig:objective} we show the learned rotation equivariance. Firstly, we observe that the equivariance decreases with the depth, up to the final part after global average pooling (GAP), regardless of the task. For the tasks where the equivariance or invariance is imposed in the task, we see an increase in the final part, which suggests that GAP plays a significant role in achieving equivariance. Secondly, we do not see any significant differences between \textit{MNIST6}, \textit{MNIST6-Rot-Inv} and \textit{MNIST6-Rot-Eq}, up to a later stage, which may indicate that early convolutional layers learn features with some amount of rotation equivariance regardless of the rotation invariance of the task. Finally, we observe that rotation equivariance is much larger then rotation invariance in the early and middle layers, which shows that CNNs do learn more rotated versions of the same feature in different channel rather then learning invariant, symmetrical features in a single channel. We conclude that introducing equivariance into the task does not significantly affect the learned equivariance of intermediate features.

\subsection{Data augmentations}
\label{sec:data_augmentation}

\begin{figure*}[t]
    \centering
    \begin{subfigure}[b]{0.48\textwidth}
        \includegraphics[width=1.0\linewidth]{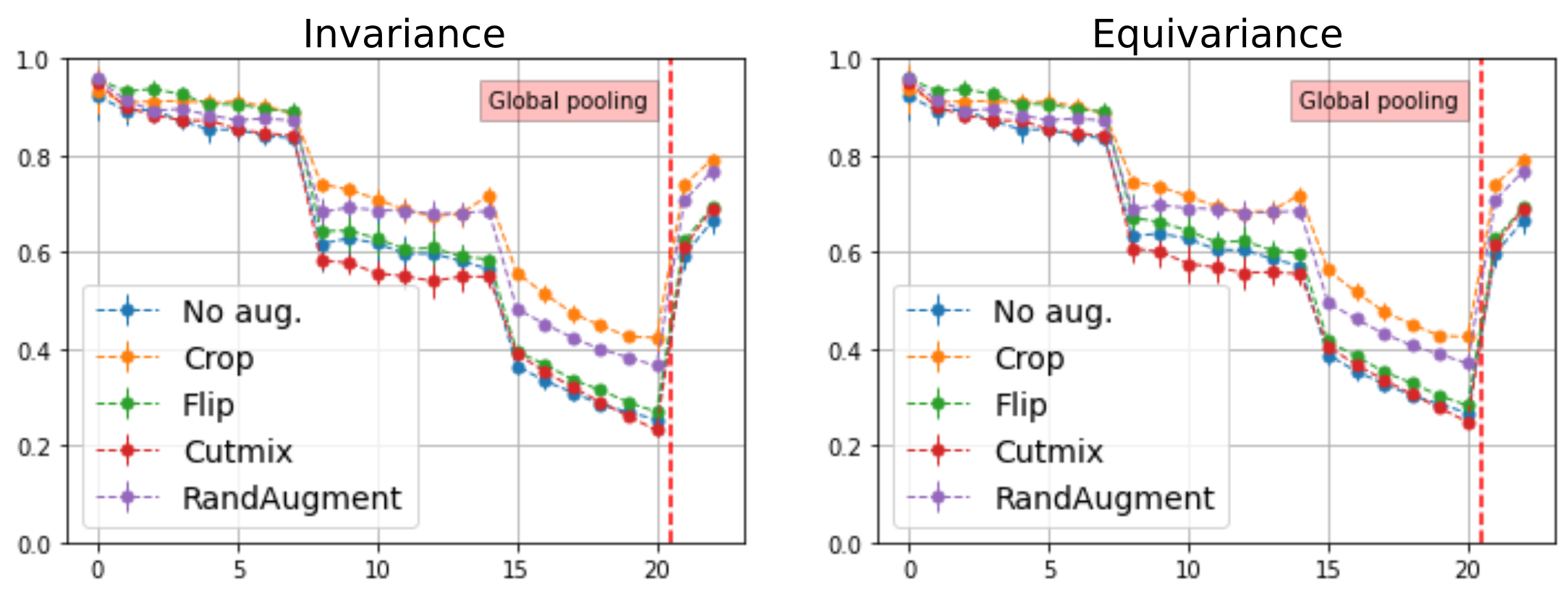}
        \caption{Data augmentations}
        \label{fig:aug_eq}
    \end{subfigure}
    \hfill
    \begin{subfigure}[b]{0.48\textwidth}
        \includegraphics[width=1.0\linewidth]{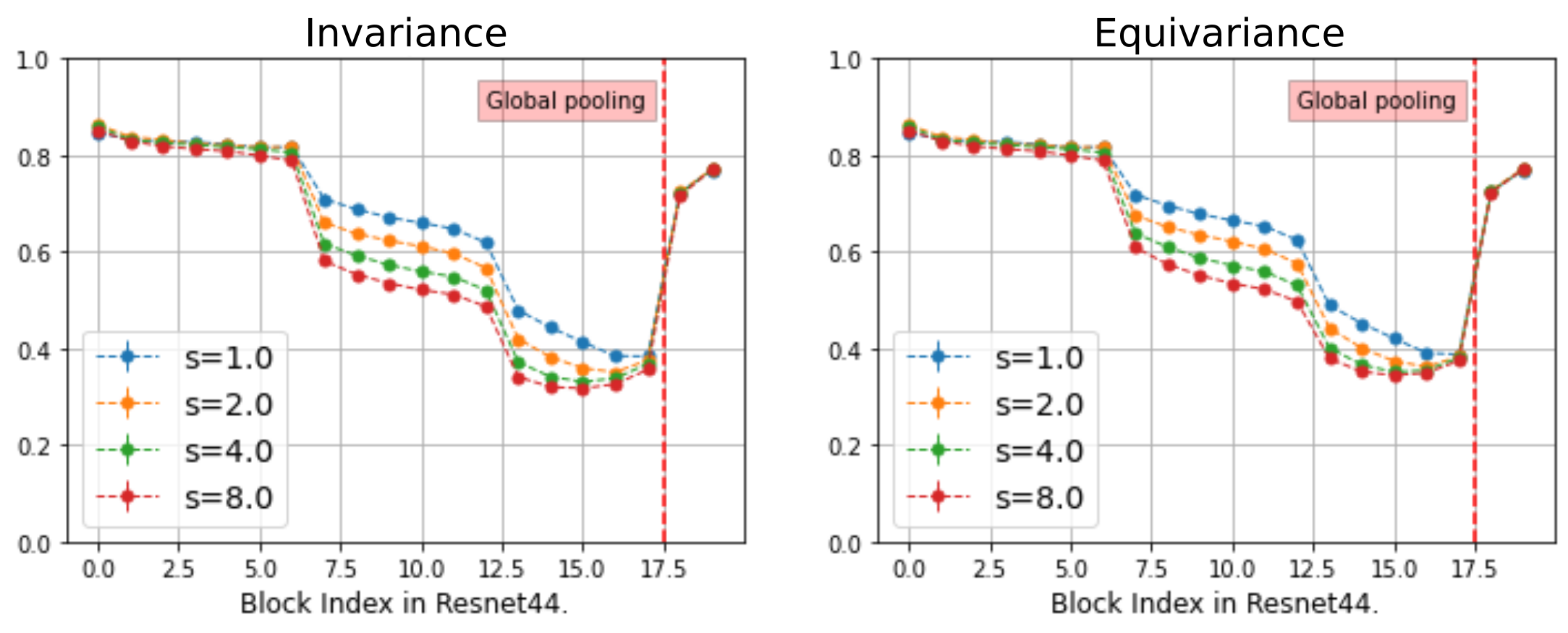}
        \caption{Model capacity: number of channels $s$}
        \label{fig:wide_eq}
    \end{subfigure}
    \caption{Measuring learned translation equivariance for (a) data augmentations and (b) model capacity. For data augmentations (a), random crops and RandAugment increase channel equivariance the most, while other strategies have no discernible improvements. For model capacity (b), smaller models learn more in- and equivariance, although the amount of in- and equivariance in the end is similar.}
\end{figure*}

\begin{figure*}[t]
    \centering
    \begin{subfigure}[b]{0.48\textwidth}
        \includegraphics[width=1.0\linewidth]{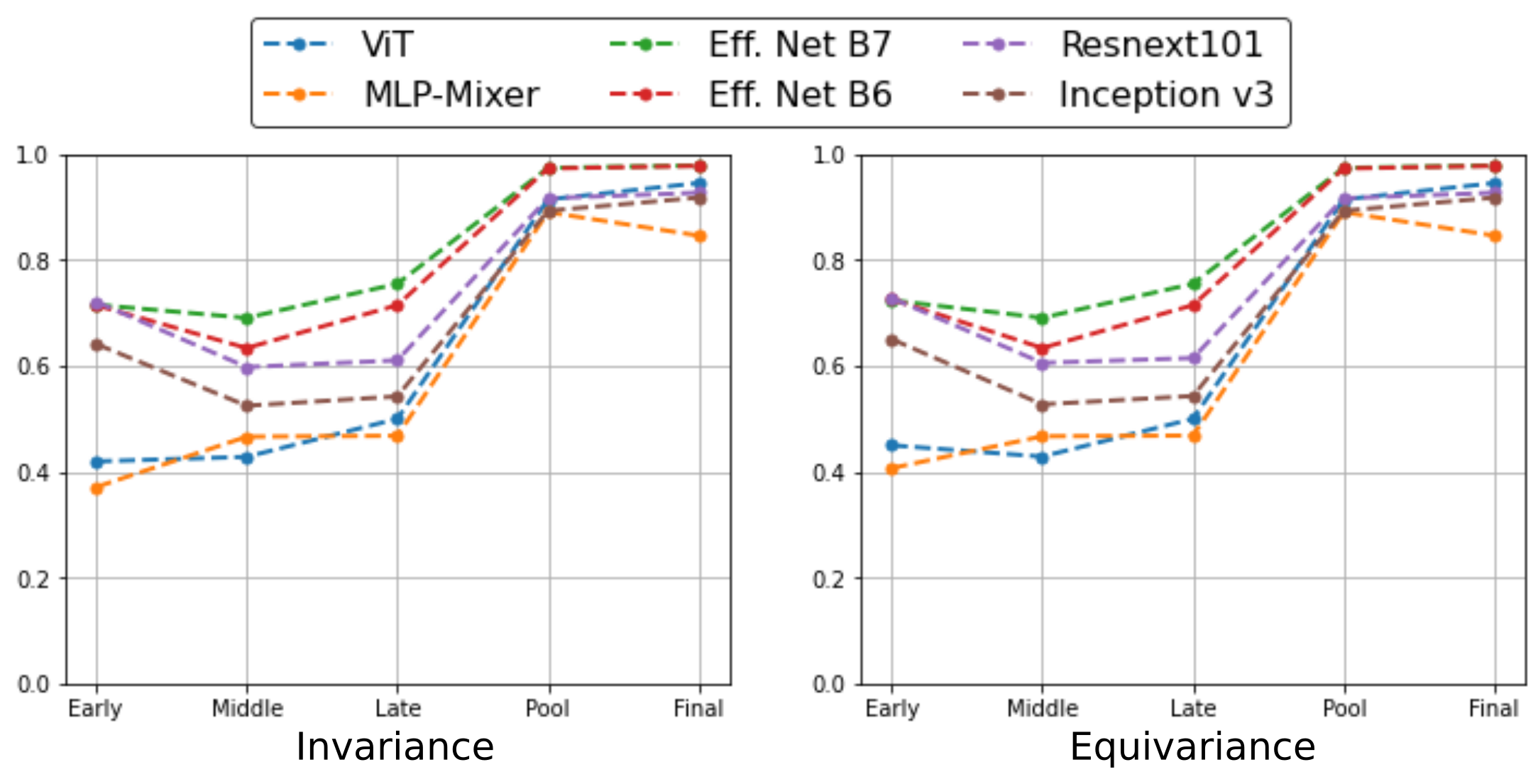}
        \caption{Learned translation equivariance}
        \label{fig:sem_trans}
    \end{subfigure}
    \hfill
    \begin{subfigure}[b]{0.48\textwidth}
        \centering
        \includegraphics[width=1.0\linewidth]{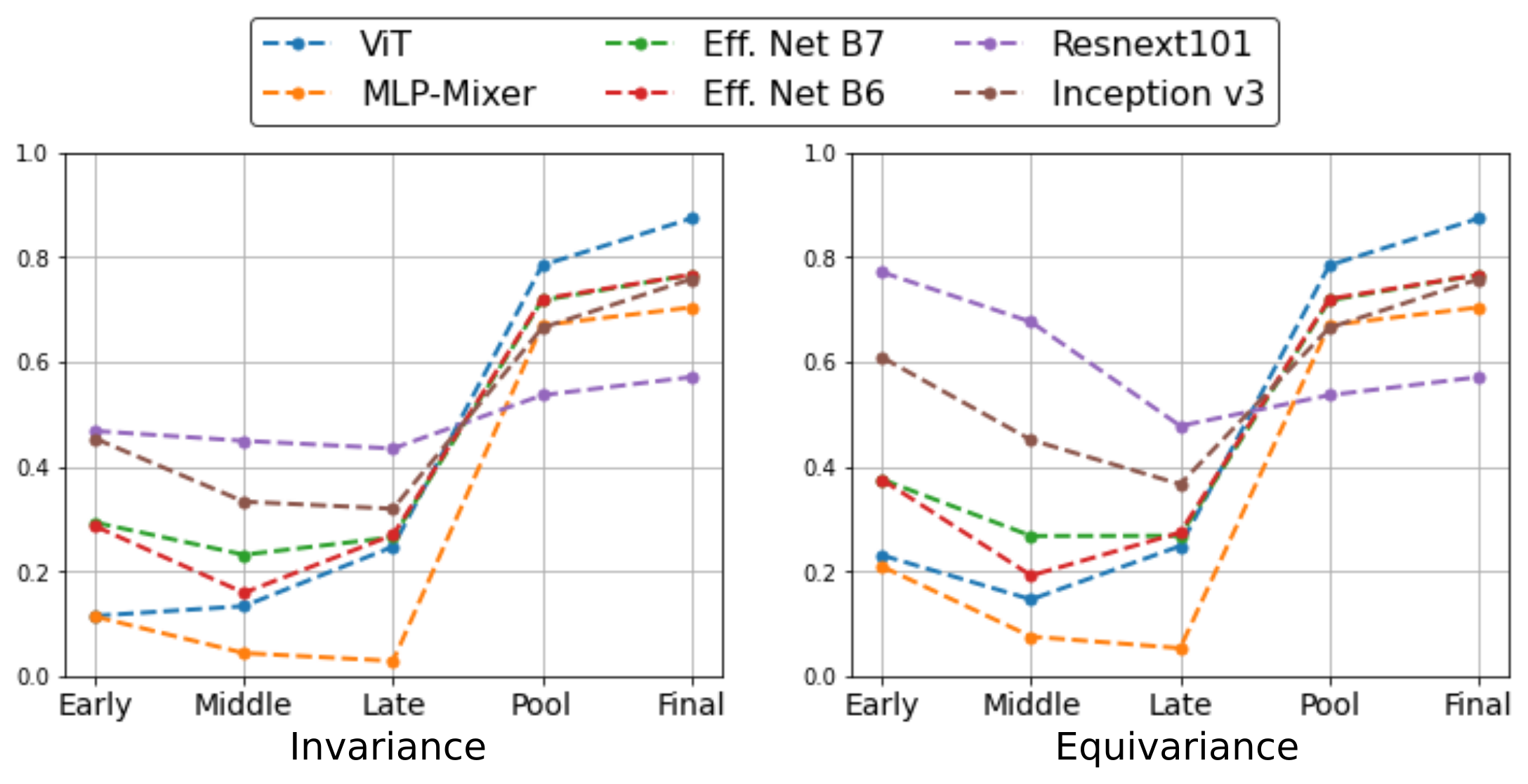}
        \caption{Learned rotation equivariance}
        \label{fig:sem_rot}
    \end{subfigure}
    \caption{Measuring learned equivariance for inductive biases. For translation (a), the CNN variants exhibit more equivariance in the intermediate representation then the Vision Transformer variants. Global pooling seems to play an important role in achieving invariance. For rotation (b), the CNN variants exhibit more equivariance in the intermediate representation then the Vision Transformer variants. The \textit{Early} and \textit{Middle} parts have more equivariance than invariance.}
\end{figure*}

By duplicating input samples under some transformation, data augmentation can induce invariance in the neural network. We study what the effect of data augmentation is on learned equivariance in intermediate representations: does data augmentation result in more invariant or equivariant intermediate features? For example, we know that the random crops data augmentation method, which essentially introduces random translations into the data, increases the model performance and translation invariance at the last layer \cite{quantifying_invariance}. The question is whether random crops increase the learned equivariance of intermediate features as well. In this experiment we study how translation equivariance is affected by different data augmentations.

We train ResNet-44\cite{resnet}, adapted for CIFAR-10\cite{cifar10}, on the CIFAR-10 dataset using one of the following augmentations: random crops, horizontal flips, CutMix \cite{cutmix}, RandAugment \cite{rand_augment}. In each experiment we compute equivariance of the trained model over 2000 images from the validation set and average the score over diagonal shifts from one to 16 pixels. Each experiment is repeated three times by training the network different random seeds. We train for 200 epochs using SGD with a batch size of 128 and a learning rate of 0.1 and a momentum of 0.9, L2 regularization at 0.0001 and weight decay at epochs 100 and 150 with a factor of 10.0.

In principle we expect the in- and equivariance to be the same since translation equivariance should be provided by the convolution. However, we include equivariance in our experiments since there are works showing that the information about location can be encoded in different channels \cite{pos_in_channel, kayhan_translation_2020}.


In Figure \ref{fig:aug_eq} we show learned translation equivariance for the tested data augmentations. Random crops and RandAugment increase the equivariance of learned features in the \textit{Middle}, \textit{Late} and \textit{Final} parts, while the other data augmentation methods do not have any significant effect, with CutMix even having less equivariance than the baseline in the \textit{Middle} part. We complement the finding of \cite{quantifying_invariance} by showing that random crops increase not only translation invariance but also translation equivariance in the intermediate layers. Also, we do not see any difference between invariance and equivariance for any data augmentation, which means that any equivariance learned is just invariance.

\subsection{Model capacity}
\label{sec:capacity}

We hypothesize that a smaller model in principle benefits from a more efficient representation and hence may learn more equivariant features. We therefore study whether model capacity influences learning translation equivariant representations. We train WideResNet-40 (WRN-40) \cite{wideResNet} models, where we scale the number of channels (the "width") in each layer by a factor $s \in \{1, 2, 4, 8\}$. We train on the CIFAR-10 dataset and measure learned translation equivariance. The hyperparameters used for training are the same as in the data augmentation experiment of Sec~\ref{sec:data_augmentation}.

In Figure \ref{fig:wide_eq} we show learned translation equivariance for different model capacities. We observe that the amount of translation equivariance is lower for the wider models, even though the amount of invariance in the final part is the same, which matches our hypothesis: an efficient representation learns to be equivariant.

\subsection{Architectures}
\label{sec:sota}

The architecture of a neural network determines which biases can be learned in training.
Vision Transformers (ViTs) \cite{vit} lack certain inductive biases present in CNNs, which has been linked to their reduced data efficiency \cite{xu2021vitae, d2021convit}. We are interested to what extent the difference in inductive bias between CNNs and Vision Transformers (ViTs) affects learned equivariance.


We test architectures as they were designed for the ImageNet dataset \cite{ImageNet}, to faithfully represent their intended inductive bias. We use the same architectures and pre-trained model weights as tested in Sec.~\ref{sec:correlation_acc}: four CNNs (EfficientNet-B6 \& EfficientNet-B7 \cite{efficient_net}, ResNeXT-101 \cite{resnext} and Inception-V3 \cite{inception_v3}) and two Vision Transformer variants (Vision Transformer \cite{vit} and MLP-Mixer \cite{mlpmixer}).
We measure both translation and rotation equivariance on trained models for 2000 images from the ImageNet validation set. We also use the same depth-wise partitioning of feature maps into parts as used in Sec.~\ref{sec:correlation_acc}. We measure translation equivariance over diagonal shifts of size 1 to 32 and rotation equivariance for ${90°, 180°, 270°}$ rotations.

In Figure \ref{fig:sem_trans} we present the results for learned translation equivariance. We can see that ViT and MLP-Mixer have less translation equivariance than CNNs in \textit{Early} and \textit{Middle} layers. This is not unexpected, as convolutions directly integrate translation equivariance, whereas Vision Transformers have to learn position embeddings that are translation equivariant.
This reduced translation equivariance could be the reason for the poor data efficiency of ViT and MLP-Mixer \cite{vit,mlpmixer} since translation equivariance improves data efficiency \cite{kayhan_translation_2020}.
Finally, we note that learned invariance and equivariance are identical for the tested models, meaning that these networks do not learn to represent different translations in different channels.


In Figure \ref{fig:sem_rot} we present the results for learned rotation equivariance. We observe that the ViT and MLP-Mixer have lower rotation equivariance than the CNNs in intermediate features, while after the GAP layer the ViT exhibits the most rotation equivariance out of all the models.
Secondly, we note that early parts of all networks learn equivariant features that are not invariant, more so than in late parts of the networks.
In contrast to the results for translation equivariance, we see that models with low rotation equivariance throughout \textit{Early}, \textit{Middle} and \textit{Late} parts (ViT, Efficient-Net B6/B7) have the highest rotation equivariance in the \textit{Final} part, while the models with highest equivariance in \textit{Early}, \textit{Middle} and \textit{Late} parts (ResNeXT-101, Inception-v3) have the least equivariance in the \textit{Final} part. This shows that high learned equivariance in the final model representation does not imply that intermediate representations are also highly equivariant.

\section{Conclusion}
We conduct a quantitative study on learned equivariance in intermediate features of CNNs and Vision Transformers trained for image recognition, using an improved measure of equivariance. We find evidence that translation equivariance in intermediate representations correlates with ImageNet validation accuracy. We show that data augmentations and reduced model capacity can increase learned equivariance in intermediate features. Also, the CNNs we test learn more translation and rotation equivariance in intermediate features than the ViTs we test.


\textbf{Limitations.} Our method allows to measure equivariance w.r.t. affine transformations only. The reason for that is the transformation $g$ with respect to which we measure the equivariance has to be a map from and to an identical discrete domain, e.g. feature maps. This restriction disqualifies continuous transformations such as rotations with any other resolution than 90 degrees, or scaling with non-integer scaling factors.


\textbf{Future work.} Learned equivariance benefits image recognition models. However, applying equivariant priors usually adds additional cost in terms of memory or computation. Future work could study whether one can apply equivariant priors selectively within a neural network, saving computing cost where networks already learn to be equivariant.
Additionally, we show that Vision Transformers learn less translation equivariance than CNNs. Future work could explore methods to increase translation invariance in Vision Transformers, to aid in their data efficiency.

\section*{Acknowledgements}

Robert-Jan Bruintjes and Jan van Gemert are financed by the Dutch Research Council (NWO) (project VI.Vidi.192.100). All authors sincerely thank everyone involved in funding this work.

{\small
\bibliographystyle{ieee_fullname}
\bibliography{main}
}






\end{document}